\documentclass{article} 
\usepackage{iclr2017_conference,times}
\usepackage{hyperref}
\usepackage{url}
\usepackage{graphicx}
\usepackage{caption}
\usepackage{subcaption}
\usepackage{tikz}
\usetikzlibrary{shapes.misc,arrows,positioning,shapes.geometric}

\newcommand{\SotA}{state-of-the-art }

\title{Deep Convolutional Neural Network Design Patterns}

\author{Leslie N.~Smith\\
	U.S. Naval Research Laboratory, Code 5514\\
	4555 Overlook Ave., SW., Washington, D.C.  20375\\
	{\tt\small leslie.smith@nrl.navy.mil}
	\And
	Nicholay Topin \\
	University of Maryland, Baltimore County \\
	1000 Hilltop Circle, 
	Baltimore, MD 21250 \\
	\texttt{ntopin1@umbc.edu} \\
}

\begin{document}

	\maketitle
	
	\begin{abstract}
		Recent research in the deep learning field has produced a plethora of new architectures.
		At the same time, a growing number of groups are applying deep learning to new applications.
		Some of these groups are likely to be composed of inexperienced deep learning practitioners who are baffled by the dizzying array of architecture choices and therefore opt to use an older architecture (i.e., Alexnet).
		Here we attempt to bridge this gap by mining the collective knowledge contained in recent deep learning research to discover underlying principles for designing neural network architectures.
		In addition, we describe several architectural innovations, including Fractal of FractalNet network, Stagewise Boosting Networks, and Taylor Series Networks (our Caffe code and prototxt files is available at https://github.com/iPhysicist/CNNDesignPatterns).
		We hope others are inspired to build on our preliminary work.
	\end{abstract}
	
	\section{Introduction}
	
	Many recent articles discuss new architectures for neural networking, especially regarding Residual Networks (\cite{he2015deep,he2016identity,larsson2016fractalnet,zhang2016residual,huang2016deep}).
	Although the literature covers a wide range of network architectures, we take a high-level view of the architectures as the basis for discovering universal principles of the design of network architecture. 
	We discuss 14 original design patterns that could benefit inexperienced practitioners who seek to incorporate deep learning in various new applications. 
	This paper addresses the current lack of guidance on design, a deficiency that may prompt the novice to rely on standard architecture, e.g., Alexnet, regardless of the architecture's suitability to the application at hand. 
	
	This abundance of research is also an opportunity to  determine which elements provide benefits in what specific contexts.
	We ask: Do universal principles of deep network design exist?
	Can these principles be mined from the collective knowledge on deep learning?
	Which architectural choices work best in any given context?
	Which architectures or parts of architectures seem elegant?
	
	Design patterns were first described by Christopher Alexander (\cite{alexander1979timeless}) in regards to the architectures of buildings and towns.
	Alexander wrote of a timeless quality in architecture that ``lives'' and this quality is enabled by building based on universal principles.
	The basis of design patterns is that they resolve a conflict of forces in a given context and lead to an equilibrium analogous to the ecological balance in nature.
	Design patterns are both highly specific, making them clear to follow, and flexible so they can be adapted to different environments and situations.
	Inspired by Alexander's work, the ``gang of four'' (\cite{gamma1995design}) applied the concept of design patterns to the architecture of object-oriented software.
	This classic computer science book describes 23 patterns that resolve issues prevalent in software design, such as ``requirements always change''.
	We were inspired by these previous works on architectures to articulate possible design patterns for convolutional neural network (CNN) architectures.
	
	Design patterns provide universal guiding principles, and here we take the first steps to defining network design patterns.
	Overall, it is an enormous task to define design principles for all neural networks and all applications, so we limit this paper to CNNs and their canonical application of image classification.
	However, we recognize that architectures must depend upon the application by defining our first design pattern; \emph{Design Pattern 1: Architectural Structure follows the Application} (we leave the details of this pattern to future work).
	In addition, these principles allowed us to discover some gaps in the existing research and  to articulate novel networks (i.e, Fractal of FractalNets, Stagewise Boosting and Taylor Series networks) and units (i.e., freeze-drop-path).
	We hope the rules of thumb articulated here are valuable for both the experienced and novice practitioners and that our preliminary work serves as a stepping stone for others to discover and share additional deep learning design patterns.
	
	\section{Related work}
	\label{sec:relwork}
	
	To the best of our knowledge, there has been little written recently to provide guidance and understanding on appropriate architectural choices\footnote{After submission we became aware of an online book being written on deep learning design patterns at http://www.deeplearningpatterns.com}.  
	The book \emph{"Neural Networks: Tricks of the Trade"} \citep{orr2003neural} contains recommendations for network models but without reference to the vast amount of research in the past few years.
	Perhaps the closest to our work is  \cite{szegedy2015rethinking} where those authors describe a few design principles based on their experiences.
	
	Much research has studied neural network architectures, but we are unaware of a recent survey of the field.
	Unfortunately, we cannot do justice to this entire body of work, so we focus on recent innovations in convolutional neural networks architectures and, in particular, on Residual Networks  \citep{he2015deep} and its recent family of variants.
	We  start with  Network In Networks \citep{lin2013network}, which describes a hierarchical network with a small network design repeatedly embedded in the overall architecture.
	\cite{szegedy2015going} incorporated this idea into their Inception architecture.
	Later, these authors proposed modifications to the original Inception design \citep{szegedy2015rethinking}.
	A similar concept was contained in the multi-scale convolution architecture \citep{liao2015competitive}.
	In the meantime, Batch Normalization \citep{ioffe2015batch} was presented as a unit within the network that makes training faster and easier.
	
	Before the introduction of Residual Networks, a few papers suggested skip connections.
	Skip connections were proposed by \cite{raiko2012deep}.
	Highway Networks \citep{srivastava2015training} use a gating mechanism to decide whether to combine the input with the layer's output and showed how these networks allowed the training of very deep networks.
	The DropIn technique \citep{smith2015gradual,smith2016gradual} also trains very deep networks by allowing a layer's input to skip the layer.
	The concept of stochastic depth via a drop-path method was introduced by  \cite{huang2016deep}.
	
	Residual Networks were introduced by \cite{he2015deep}, where the authors describe their network that won the 2015 ImageNet Challenge.
	They were able to extend the depth of a network from tens to hundreds of layers and in doing so, improve the network's performance.
	The authors followed up with another paper \citep{he2016identity} where they investigate why identity mappings help and report results for a network with more than a thousand layers.
	The research community took notice of this architecture and many modifications to the original design were soon proposed.
	
	The Inception-v4 paper \citep{szegedy2016inception} describes the impact of residual connections on their Inception architecture and compared these results with the results from an updated Inception design.
	The Resnet in Resnet paper \citep{targ2016resnet} suggests a duel stream architecture.
	\cite{veit2016residual} provided an understanding of Residual Networks as an ensemble of relatively shallow networks.
	These authors illustrated how these residual connections allow the input to follow an exponential number of paths through the architecture.
	At the same time, the FractalNet paper \citep{larsson2016fractalnet} demonstrated training deep networks with a symmetrically repeating architectural pattern.
	As described later, we found the symmetry introduced in their paper intriguing.
	In a similar vein,  Convolutional Neural Fabrics \citep{saxena2016convolutional} introduces a three dimensional network, where the usual depth through the network is the first dimension.
	
	Wide Residual Networks \citep{zagoruyko2016wide} demonstrate that simultaneously increasing both depth and width leads to improved performance.
	In Swapout \citep{singh2016swapout},  each layer can be dropped, skipped, used normally, or combined with a residual.
	Deeply Fused Nets \citep{wang2016deeply} proposes networks with multiple paths.
	In the Weighted Residual Networks paper \citep{shen2016weighted}, the authors recommend a weighting factor for the output from the convolutional layers, which  gradually introduces the trainable layers.
	Convolutional Residual Memory Networks \citep{moniz2016convolutional} proposes an architecture that combines a convolutional Residual Network with an LSTM memory mechanism.
	For Residual of Residual Networks \citep{zhang2016residual}, the authors propose adding a hierarchy of skip connections where the input can skip a layer, a module, or any number of modules.
	DenseNets \citep{huang2016densely} introduces a network where each module is densely connected; that is, the output from a layer is input to all of the other layers in the module.
	In the Multi-Residual paper \citep{abdi2016multi}, the authors propose expanding a residual block width-wise to contain multiple convolutional paths.
	Our Appendix \ref{sec:equiv} describes the close relationship between many of these Residual Network variants.
	
	\section{Design Patterns}

	We reviewed the literature specifically to extract commonalities and reduce their designs down to fundamental elements that might be considered design patterns.
	It seemed clear to us that in reviewing the literature some design choices are elegant while others are less so.
	In particular, the patterns described in this paper are the following:
	\emph{
		\begin{enumerate}
			\item Architectural Structure follows the Application
			\item Proliferate Paths
			\item Strive for Simplicity
			\item Increase Symmetry
			\item Pyramid Shape
			\item Over-train
			\item Cover the Problem Space
			\item Incremental Feature Construction
			\item Normalize Layer Inputs
			\item Input Transition
			\item Available Resources Guide  Layer Widths
			\item Summation Joining
			\item Down-sampling Transition
			\item Maxout for Competition
		\end{enumerate}
	}
	
	\subsection{High Level Architecture Design}
	\label{sec:design}
	
	Several researchers have pointed out that the winners of the ImageNet Challenge \citep{russakovsky2015imagenet} have successively used deeper networks (as seen in, \cite{krizhevsky2012imagenet}, \cite{szegedy2015going}, \cite{simonyan2014very}, \cite{he2015deep}).
	It is also apparent to us from the ImageNet Challenge that multiplying the number of paths through the network is a recent trend that is illustrated in the progression from Alexnet to Inception to ResNets.
	For example, \cite{veit2016residual} show that ResNets can be considered to be an exponential ensemble of networks with different lengths.
	\emph{Design Pattern 2: Proliferate Paths} is based on the idea that ResNets can be an exponential ensemble of networks with different lengths.
	One proliferates paths by including a multiplicity of branches in the architecture. Recent examples include FractalNet (\citealt{larsson2016fractalnet}), Xception (\citealt{chollet2016deep}), and Decision Forest Convolutional Networks (\citealt{ioannou2016decision}).
	
	Scientists have embraced simplicity/parsimony for centuries.
	Simplicity was exemplified in the paper "Striving for Simplicity" (\citealt{springenberg2014striving}) by achieving \SotA  results with fewer types of units.
	\emph{Design Pattern 3: Strive for Simplicity} suggests using fewer types of units and keeping the network as simple as possible.
	We also noted a special degree of elegance in the FractalNet (\citealt{larsson2016fractalnet}) design, which we attributed to the symmetry of its structure.
	\emph{Design Pattern 4: Increase Symmetry} is derived from the fact that architectural symmetry is typically considered a sign of beauty and quality.
	In addition to its symmetry, FractalNets also adheres to the \emph{Proliferate Paths} design pattern so we used it as the baseline of our experiments in Section \ref{sec:exp}.
	
	An essential element of design patterns is the examination of trade-offs in an effort to understand the relevant forces.
	One  fundamental trade-off is the maximization of representational power versus elimination of redundant and non-discriminating information.
	It is universal in all convolutional neural networks that the activations are downsampled and the number of channels increased from the input to the final layer, which is exemplified in Deep Pyramidal Residual Networks (\citet{han2016deep}).
	\emph{Design Pattern 5: Pyramid Shape} says there should be an overall smooth downsampling  combined with an increase in the number of channels throughout the architecture.
	
	Another  trade-off in deep learning is training accuracy versus the ability of the network to generalize to non-seen cases.
	The ability to generalize is an important virtue of deep neural networks.
	Regularization is commonly used to improve generalization, which includes methods such as dropout (\citealt{srivastava2014dropout}) and drop-path (\citealt{huang2016deep}).
	As noted by \citealt{srivastava2014understanding}, dropout improves generalization by injecting noise in the architecture.
	We believe regularization techniques and prudent noise injection  during training improves generalization (\citealt{srivastava2014understanding}, \citealt{gulcehre2016noisy}).
	\emph{Design Pattern 6: Over-train} includes any training method where the network is trained on a harder problem than necessary to improve generalization performance of inference.
	\emph{Design Pattern 7: Cover the Problem Space} with the training data is another way to improve generalization (e.g., \citealt{ratner2016data},  \citealt{hu2016frankenstein}, \citealt{wong2016understanding}, \citealt{johnson2016driving}).
	Related to regularization methods, cover the problem space includes the use of noise (\citealt{rasmus2015semi},  \citealt{krause2015unreasonable}, \citealt{pezeshki2015deconstructing}) and data augmentation, such as random cropping, flipping, and varying brightness, contrast, and the like.
	
	
	\subsection{Detailed Architecture Design}
	\label{sec:detail}
	
	A common thread throughout many of the more successful architectures is to make each layer's ``job''  easier.
	Use of very deep networks is  an example because any single layer only needs to incrementally modify the input, and this partially explains the success of Residual Networks, since in very deep networks, a  layer's output is likely similar to the input; hence adding the input to the layer's output makes the layer's job incremental.
	Also, this concept is part of the motivation behind design pattern 2 but it extends beyond that.
	\emph{Design Pattern 8: Incremental Feature Construction} recommends  using short skip lengths in ResNets.
	A recent paper (\citet{alain2016understanding}) showed in an experiment that using an identity skip length of 64 in a network of depth 128 led to the first portion of the network not being trained.
	
	\emph{Design Pattern 9: Normalize Layer Inputs} is another way to make a layer's job  easier.
	Normalization of layer inputs has been shown to improve training and accuracy but the underlying reasons are not clear (\citealt{ioffe2015batch}, \citealt{ba2016layer}, \citealt{salimans2016weight}).
	The Batch Normalization paper (\citealt{ioffe2015batch})  attributes the benefits to handling internal covariate shift, while the authors of streaming normalization (\citealt{liao2016streaming}) express that it might be otherwise.
	We feel that normalization puts all the layer's input samples on more equal footing (analogous to a units conversion scaling), which allows back-propagation to train more effectively.
	
	Some research, such as Wide ResNets (\citealt{zagoruyko2016wide}),  has shown that increasing the number of channels improves performance but there are additional costs with extra channels.
	The input data for many of the benchmark datasets  have 3 channels (i.e., RGB).
	\emph{Design Pattern 10: Input Transition} is based on the common occurrence that the output from the first layer of a CNN significantly increases the number of channels from 3.
	A few examples of this increase in channels/outputs at the first layer for ImageNet are AlexNet (96 channels), Inception (32), VGG (224), and ResNets (64).
	Intuitively it makes sense to increase the number of channels from 3 in the first layer as it allows the input data to be examined many ways but it is not clear how many outputs are best.
	Here, the trade-off is that of cost versus accuracy.
	Costs include the number of parameters in the network, which directly affects the computational and storage costs of training and inference.
	\emph{Design Pattern 11: Available Resources Guide Layer Widths} is based on balancing costs against an application's requirements.
	Choose the number of outputs of the first layer based on memory and computational resources and desired accuracy. 
	
	
	\subsubsection{Joining branches: Concatenation, summation/mean,  Maxout}
	\label{sec:joining}
	
	When there are multiple branches, three methods have been used to combine the outputs; concatenation, summation (or mean), or Maxout.
	It seems that different researchers have their favorites and there hasn't been any motivation for using one in preference to another.
	In this Section, we propose some  rules for deciding how to combine branches.
	
	Summation is one of the most common ways of combining branches.
	\emph{Design Pattern 12:  Summation Joining} is where the joining  is performed by summation/mean.
	Summation is the preferred joining mechanism for Residual Networks because it allows each branch to compute corrective terms (i.e., residuals) rather than the entire approximation.
	The difference between summation and mean (i.e., fractal-join) is best understood by considering drop-path
	(\citealt{huang2016deep}).
	In a Residual Network where the input skip connection is always present, summation causes the  layers to learn the residual (the difference from the input).
	On the other hand, in networks with several branches, where any branch can be dropped (e.g., FractalNet (\cite{larsson2016fractalnet})), using the mean is preferable as it keeps the output smooth as branches are randomly dropped.
	
	Some researchers seem to prefer concatenation (e.g., \cite{szegedy2015going}).
	\emph{Design Pattern 13: Down-sampling Transition}  recommends using concatenation joining for increasing the number of outputs when pooling.
	That is, when down-sampling by pooling or using a stride greater than 1, a good way to combine branches is to concatenate the output channels, hence smoothly accomplishing both joining and an increase in the number of channels that typically accompanies down-sampling.
	
	Maxout has been used for competition, as in locally competitive networks (\citealt{srivastava2014understanding}) and competitive multi-scale networks \cite{liao2015competitive}.
	\emph{Design Pattern 14: Maxout for Competition} is based on Maxout choosing only one of the activations, which  is in contrast to summation or mean where the activations are ``cooperating''; here, there is a ``competition'' with only one ``winner''.
	For example, when each branch is composed of different sized kernels, Maxout is useful for incorporating scale invariance in an analogous way to how max pooling enables translation invariance.

	
	\section{Experiments}
	\label{sec:exp}

	\subsection{Architectural Innovations}
	\label{sec:innovations}
	
	In elucidating these fundamental design principles, we also discovered  a few architectural innovations.
	In this section we will describe these innovations.
	
\begin{figure}[ht]
		\centering
		\begin{subfigure}[b]{0.47\textwidth}
			\begin{tikzpicture}[>=latex', semithick]
			\tikzset{
				block/.style= {draw, rectangle, fill=pink, align=center,minimum width=0.75cm,minimum height=0.25cm},
				block_wide1/.style= {draw, rectangle, fill=green, align=center,minimum width=2.5cm,minimum height=0.25cm},
				block_wide2/.style= {draw, rectangle, fill=green, align=center,minimum width=5cm,minimum height=0.25cm},
				block_wide0/.style= {draw, rectangle, fill=green, align=center,minimum width=1.5cm,minimum height=0.25cm},
				block_grey/.style= {draw, rounded rectangle, fill=lightgray, align=center,minimum width=1.5cm,minimum height=0.5cm},
				pool/.style= {draw, rectangle, fill=yellow, align=center,minimum width=0.75cm,minimum height=0.25cm},
				predict/.style= {draw, rectangle, fill=blue, align=center,minimum width=0.75cm,minimum height=0.25cm},
			}
			\node [coordinate] (source) {{\small $z$}};
			\node [block_wide2, below =4cm of source] (bottom_mean) {};
			\node [coordinate, above left =-0.5cm and 1.25cm of source] (c0top) {};
			\node [block, below =1.5cm of c0top] (c0b0) {};
			\node [block_wide1, right =0.875cm of c0b0] (mid_mean) {};
			\node [coordinate, below =1.75cm of c0b0] (c0bot) {};
			\node [coordinate, above right =-0.5cm and 1.25cm of source] (branch1) {};
			\node [coordinate, above left = -0.25cm and 0.75 of branch1] (c1top) {};
			\node [coordinate, above right = -0.25cm and 0.75 of branch1] (c2top) {};
			\node [block, below =0.5cm of c1top] (c1b0) {};
			\node [block, below =0.25cm of c2top] (c2b0) {};
			\node [block, below =0.25cm of c2b0] (c2b1) {};
			\node [coordinate, below =0.25 of mid_mean] (branch2) {};
			\node [coordinate, above left =-0.25 and 0.75 of branch2] (c1mid) {};
			\node [coordinate, above right =-0.25 and 0.75 of branch2] (c2mid) {};
			\node [block, below =0.5cm of c1mid] (c1b1) {};
			\node [block, below =0.25cm of c2mid] (c2b2) {};
			\node [block, below =0.25cm of c2b2] (c2b3) {};
			\node [coordinate, below = 0.25 of c2b1] (c2b1_node) {};
			\node [coordinate, below = 0.25 of c2b3] (c2b3_node) {};
			\node [coordinate, below = 0.5 of c1b0] (c1b0_node) {};
			\node [coordinate, below = 0.5 of c1b1] (c1b1_node) {};
			\node [coordinate, below = 0.75 of bottom_mean] (bottom_node) {};
			
			\node [block, above left = -1 and 1 of bottom_node] (conv_layer) {};
			\node [text width = 3cm, right = 1 of conv_layer.center] (conv_label) {convolutional layer};
			\node [pool, below = 0.25cm of conv_layer] (pool_layer) {};
			\node [text width = 3cm, right = 1 of pool_layer.center] (pool_label) {pooling layer};
			\node [predict, below = 0.25 of pool_layer] (predict_layer) {};
			\node [text width = 3cm, right = 1 of predict_layer.center] (predict_label) {prediction layer};
			\node [block_wide0, below = 0.25 of predict_layer] (average_layer) {};
			\node [text width = 3cm, right = 1 of average_layer.center] (average_label) {joining layer};
			\node [block_grey, below = 0.25 of average_layer] (fractal_block) {};
			\node [text width = 3cm, right = 1 of fractal_block.center] (fractal_label) {FractalNet module};
			
			\path[->] 	(c0top.south) edge [out = 270, in = 90, looseness = 1] (c0b0.north)
			(c0b0.south) edge [out = 270, in = 90, looseness = 1] (c0bot.north)
			(c1top.south) edge [out = 270, in = 90, looseness = 1] (c1b0.north)
			(c2top.south) edge [out = 270, in = 90, looseness = 1] (c2b0.north)
			(c2b0.south) edge [out = 270, in = 90, looseness = 1] (c2b1.north)
			(c1mid.south) edge [out = 270, in = 90, looseness = 1] (c1b1.north)
			(c2mid.south) edge [out = 270, in = 90, looseness = 1] (c2b2.north)
			(c2b2.south) edge [out = 270, in = 90, looseness = 1] (c2b3.north)
			(c2b1.south) edge [out = 270, in = 90, looseness = 1] (c2b1_node.north)
			(c2b3.south) edge [out = 270, in = 90, looseness = 1] (c2b3_node.north)
			(c1b0.south) edge [out = 270, in = 90, looseness = 1] (c1b0_node.north)
			(c1b1.south) edge [out = 270, in = 90, looseness = 1] (c1b1_node.north)
			(bottom_mean.south) edge [out = 270, in = 90, looseness = 1] (bottom_node.north)
			;
			\path[-]	(source.south) edge [out = 270, in = 90, looseness = 1] (c0top.north)
			(source.south) edge [out = 270, in = 90, looseness = 1] (branch1.north)
			(branch1.south) edge [out = 270, in = 90, looseness = 1] (c1top.north)
			(branch1.south) edge [out = 270, in = 90, looseness = 1] (c2top.north)
			(branch2.south) edge [out = 270, in = 90, looseness = 1] (c1mid.north)
			(branch2.south) edge [out = 270, in = 90, looseness = 1] (c2mid.north)
			(mid_mean.south) edge [out = 270, in = 90, looseness = 1] (branch2.north)
			;
			\end{tikzpicture}
			\caption{FractalNet module}
			\label{fig:Fractmodule}
		\end{subfigure}
		\quad
		\centering
		\begin{subfigure}[b]{0.47\textwidth}
			\begin{tikzpicture}[>=latex', semithick]
			\tikzset{
				block/.style= {draw, rounded rectangle, fill=lightgray, align=center,minimum width=1.5cm,minimum height=0.5cm},
				pool/.style= {draw, rectangle, fill=yellow, align=center,minimum width=0.75cm,minimum height=0.25cm},
				predict/.style= {draw, rectangle, fill=blue, align=center,minimum width=0.75cm,minimum height=0.25cm},
				block_wide1/.style= {draw, rectangle, fill=green, align=center,minimum width=2.5cm,minimum height=0.25cm},
				block_wide2/.style= {draw, rectangle, fill=green, align=center,minimum width=5cm,minimum height=0.25cm}
			}
			\node [coordinate] (source) {{\small $z$}};
			\node [block_wide2, below =7.25cm of source] (bottom_mean) {};
			\node [coordinate, above left =-0.5cm and 1.25cm of source] (c0top) {};
			\node [block, below =3cm of c0top] (c0b0) {};
			\node [pool, below=0.25 of c0b0] (c0p0) {};
			\node [pool, below=0.25 of c0p0] (c0p1) {};
			\node [pool, below=0.25 of c0p1] (c0p2) {};
			\node [pool, below=0.25 of c0p2] (c0p3) {};
			\node [coordinate, below =1.7cm of c0p2] (c0bot) {};
			\node [coordinate, above right =-0.5cm and 1.25cm of source] (branch1) {};
			\node [block_wide1, below =3.125cm of branch1] (mid_mean) {};
			\node [coordinate, above left = -0.25cm and 0.75 of branch1] (c1top) {};
			\node [coordinate, above right = -0.25cm and 0.75 of branch1] (c2top) {};
			\node [block, below =0.75cm of c1top] (c1b0) {};
			\node [pool, below=0.25 of c1b0] (c1p0) {};
			\node [pool, below=0.25 of c1p0] (c1p1) {};
			\node [block, below =0.25cm of c2top] (c2b0) {};
			\node [pool, below=0.25 of c2b0] (c2p0) {};
			\node [block, below =0.25cm of c2p0] (c2b1) {};
			\node [pool, below=0.25 of c2b1] (c2p1) {};
			\node [coordinate, below =0.25 of mid_mean] (branch2) {};
			\node [coordinate, above left =-0.25 and 0.75 of branch2] (c1mid) {};
			\node [coordinate, above right =-0.25 and 0.75 of branch2] (c2mid) {};
			\node [block, below =0.75cm of c1mid] (c1b1) {};
			\node [pool, below=0.25 of c1b1] (c1p2) {};
			\node [pool, below=0.25 of c1p2] (c1p3) {};
			\node [block, below =0.25cm of c2mid] (c2b2) {};
			\node [pool, below=0.25 of c2b2] (c2p2) {};
			\node [block, below =0.25cm of c2p2] (c2b3) {};
			\node [pool, below=0.25 of c2b3] (c2p3) {};
			\node [coordinate, below = 0.3 of c2p1] (c2p1_node) {};
			\node [coordinate, below = 0.3 of c2p3] (c2p3_node) {};
			\node [coordinate, below = 0.55 of c1p1] (c1p0_node) {};
			\node [coordinate, below = 0.55 of c1p3] (c1p3_node) {};
			\node [pool, below=0.25 of bottom_mean] (bottom_pool) {};
			\node [predict, below=0.25 of bottom_pool] (bottom_predict) {};
			\node [coordinate, below = 0.5 of bottom_predict] (bottom_node) {};
			
			\path[->] 	(c0top.south) edge [out = 270, in = 90, looseness = 1] (c0b0.north)
			(c0b0.south) edge [out = 270, in = 90, looseness = 1] (c0p0.north)
			(c0p0.south) edge [out = 270, in = 90, looseness = 1] (c0p1.north)
			(c0p1.south) edge [out = 270, in = 90, looseness = 1] (c0p2.north)
			(c0p2.south) edge [out = 270, in = 90, looseness = 1] (c0p3.north)
			(c0p3.south) edge [out = 270, in = 90, looseness = 1] (c0bot.north)
			(c1top.south) edge [out = 270, in = 90, looseness = 1] (c1b0.north)
			(c2top.south) edge [out = 270, in = 90, looseness = 1] (c2b0.north)
			(c2b0.south) edge [out = 270, in = 90, looseness = 1] (c2p0.north)
			(c2p0.south) edge [out = 270, in = 90, looseness = 1] (c2b1.north)
			(c2b1.south) edge [out = 270, in = 90, looseness = 1] (c2p1.north)
			(c1mid.south) edge [out = 270, in = 90, looseness = 1] (c1b1.north)
			(c2mid.south) edge [out = 270, in = 90, looseness = 1] (c2b2.north)
			(c2b2.south) edge [out = 270, in = 90, looseness = 1] (c2p2.north)
			(c2p2.south) edge [out = 270, in = 90, looseness = 1] (c2b3.north)
			(c2b3.south) edge [out = 270, in = 90, looseness = 1] (c2p3.north)
			(c2p1.south) edge [out = 270, in = 90, looseness = 1] (c2p1_node.north)
			(c2p3.south) edge [out = 270, in = 90, looseness = 1] (c2p3_node.north)
			(c1b0.south) edge [out = 270, in = 90, looseness = 1] (c1p0.north)
			(c1p0.south) edge [out = 270, in = 90, looseness = 1] (c1p1.north)
			(c1p1.south) edge [out = 270, in = 90, looseness = 1] (c1p0_node.north)
			(c1b1.south) edge [out = 270, in = 90, looseness = 1] (c1p2.north)
			(c1p2.south) edge [out = 270, in = 90, looseness = 1] (c1p3.north)
			(c1p3.south) edge [out = 270, in = 90, looseness = 1] (c1p3_node.north)
			(bottom_mean.south) edge [out = 270, in = 90, looseness = 1] (bottom_pool.north)
			(bottom_pool.south) edge [out = 270, in = 90, looseness = 1] (bottom_predict.north)
			(bottom_predict.south) edge [out = 270, in = 90, looseness = 1] (bottom_node.north)
			;
			\path[-]	(source.south) edge [out = 270, in = 90, looseness = 1] (c0top.north)
			(source.south) edge [out = 270, in = 90, looseness = 1] (branch1.north)
			(branch1.south) edge [out = 270, in = 90, looseness = 1] (c1top.north)
			(branch1.south) edge [out = 270, in = 90, looseness = 1] (c2top.north)
			(branch2.south) edge [out = 270, in = 90, looseness = 1] (c1mid.north)
			(branch2.south) edge [out = 270, in = 90, looseness = 1] (c2mid.north)
			(mid_mean.south) edge [out = 270, in = 90, looseness = 1] (branch2.north)
			;
			\end{tikzpicture}
			\caption{Fractal of FractalNet architecture}
			\label{fig:FoFarch}
		\end{subfigure}
		\caption{(a) The FractalNet module and (b) the FoF architecture.}
		\label{fig:fractalArch}
	\end{figure}
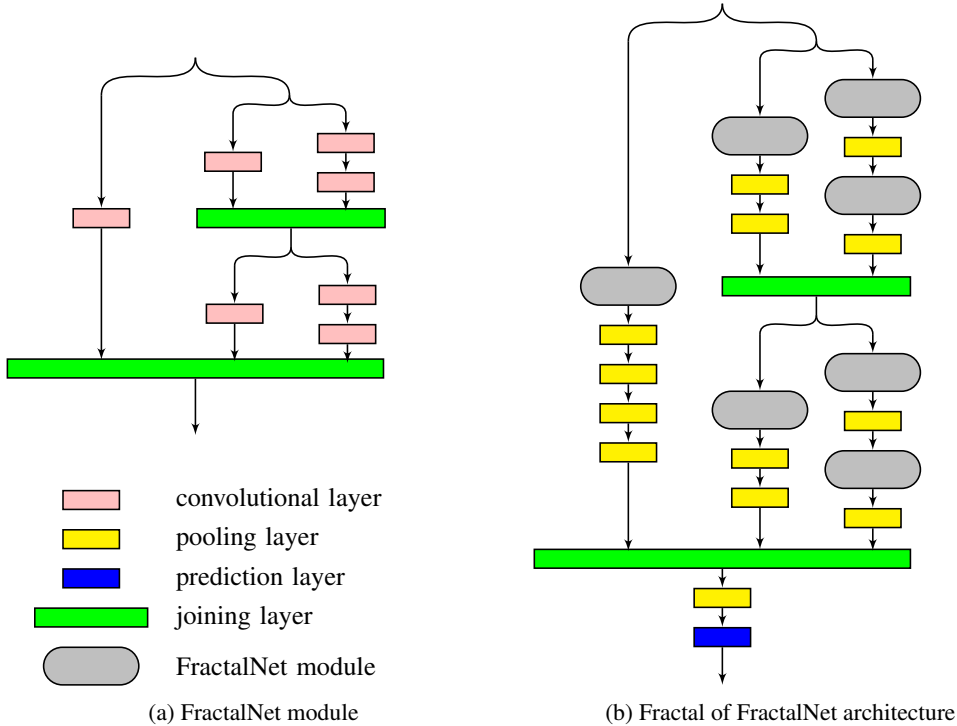

	First, we recommended combining summation/mean, concatenation, and maxout joining mechanisms with differing roles within a single architecture, rather than the typical situation where only one is used throughout.
	Next, \emph{Design Pattern 2: Proliferate Branches} led us to modify the overall sequential pattern of modules in the FractalNet architecture.
	Instead of lining up the modules for maximum depth, we arranged the modules in a fractal pattern as shown in \ref{fig:FoFarch}, which we named  a Fractal of FractalNet (FoF) network, where we exchange depth for a greater number of paths.

	
	\subsubsection{Freeze-drop-path and Stagewise Boosting Networks (SBN)}
	\label{sec:freeze}
	
	Drop-path was introduced by \cite{huang2016deep}, which works by randomly removing  branches during an iteration of training, as though that path doesn't exist in the network.
	Symmetry considerations led us to an opposite method that we named freeze-path.
	Instead of removing a branch from the network during training, we freeze the weights, as though the learning rate was set to zero.
	A similar idea has been proposed for recurrent neural networks (\citealt{krueger2016zoneout}).
	
	The potential usefulness of combining drop-path and freeze-path, which we named freeze-drop-path, is best explained in the non-stochastic case.
	Figure \ref{fig:fractalArch} shows an example of a fractal of FractalNet architecture. 
	Let's say we start training only the leftmost branch in Figure \ref{fig:FoFarch} and use drop-path on  all of the other branches.
	This branch should train quickly since it has only a relatively few parameters compared to the entire network.
	Next we freeze the weights in that branch and allow the next branch to the right to be active.
	If the leftmost branch is providing a good function approximation,  the next branch works to produce a ``small'' corrective term.
	Since the next branch contains more layers than the previous branch and the corrective term should be easier to approximate than the original function, the network should attain greater accuracy.
	One can continue this process from left to right to train the entire network.
	We used freeze-drop-path as the final/bottom join in the FoF architecture in Figure \ref{fig:FoFarch}  and  named this the Stagewise Boosting Networks (SBN) because they are analogous to stagewise boosting (\citealt{friedman2001elements}). 
	The idea of boosting neural networks is not new (\citealt{schwenk2000boosting}) but this architecture is new.
	In Appendix \ref{sec:impdetails} we  discuss the implementation we tested.
	
	\subsubsection{Taylor Series Networks (TSN)}
	\label{sec:TSN}
	
	Taylor series expansions are classic and well known as a function approximation method, which is:
	\begin{equation}
	f(x+h) = f(x) + h f'(x)  + h^2 f''(x)/2 + ... 
	\end{equation}
	Since neural networks are also function approximators, it is a short leap from FoFs and SBNs to consider the branches of that network as terms in a Taylor series expansion.
	Hence, the Taylor series implies  squaring the second branch before the summation joining unit, analogous to the second order term in the  expansion.
	Similarly, the third branch would be cubed.
	We call this ``Taylor Series Networks'' (TSN) and there is precedence for this idea in the literature with polynomial networks (\citealt{livni2014computational}) and multiplication in networks (e.g. \citealt{lin2015bilinear}. 
	The implementation details of this TSN are discussed in the Appendix.

	\begin{table}[tbh]
		\begin{center}
			\caption{Comparison of test accuracy results  at the end of the training.}
			\begin{tabular}{| c | c | c |}
				\hline
				Architecture  & CIFAR-10 (\%) & CIFAR-100 (\%)   \\ \hline \hline
				FractalNet &  $ 93.4 $  & $ 72.5 $  \\ \hline
				FractalNet + Concat &  $ 93.0 $  & $ 72.6 $  \\ \hline
				FractalNet + Maxout &  $ 91.8  $  & $ 70.2 $ \\ \hline
				FractalNet + Avg pooling &  $ 94.3 $  & $ 73.4 $  \\ \hline
				FoF &  $ 92.6 $  & $ 73.1 $  \\ \hline
				SBN &  $ 91.4 $  & $ 68.7 $ \\ \hline
				TSN &  $ 90.7 $  & $ 68.4 $  \\ \hline
			\end{tabular}
			\vspace{5pt}
			\label{tab:results}
		\end{center}
		\vspace{-10pt}
	\end{table}
	
	
	\begin{figure} [tbh]
		\centering
		\begin{subfigure}[b]{0.47\textwidth}
			\includegraphics[width=\textwidth]{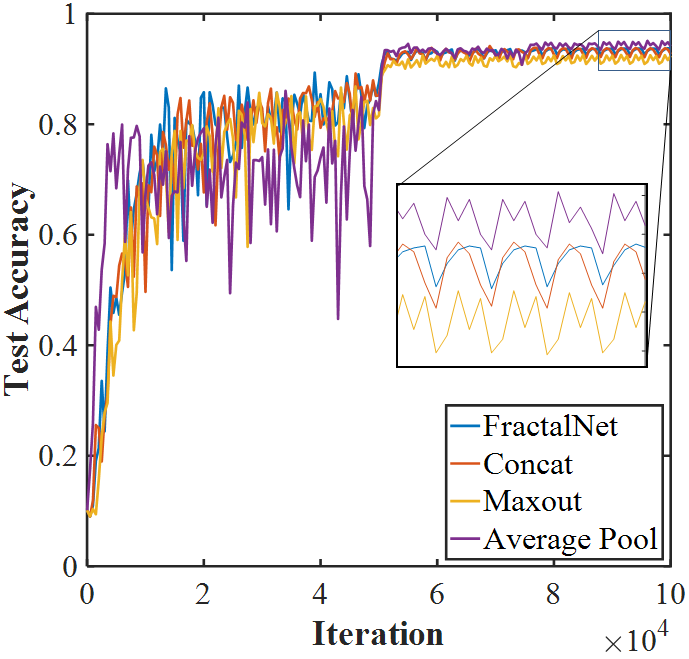}
			\caption{Cifar-10}
			\label{fig:Cifar10fractal}
		\end{subfigure}
		\quad
		\centering
		\begin{subfigure}[b]{0.47\textwidth}
			\includegraphics[width=\textwidth]{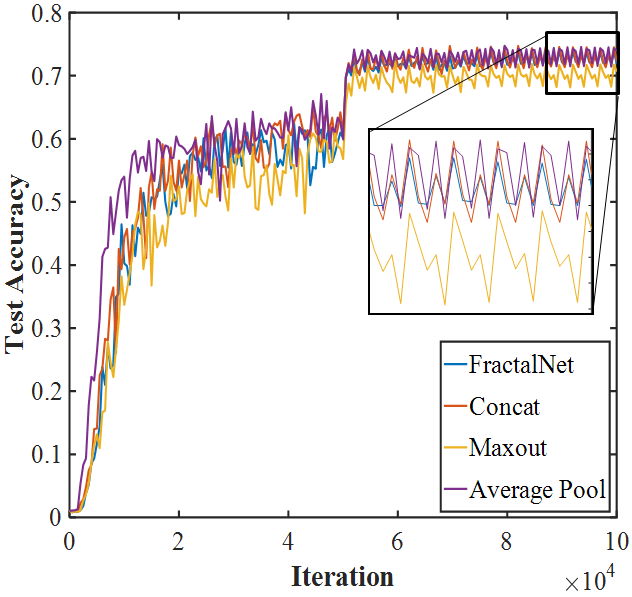}
			\caption{Cifar-100}
			\label{fig:Cifar100fractal}
		\end{subfigure}
		\caption{Test accuracy of the original FractalNet compared with replacing some of the fractal-joins with Concatenation or Maxout, and  when replacing max pooling with average pooling.}
		\label{fig:fractalCompare}
	\end{figure}
	
	\begin{figure} [tbh]
		\centering
		\begin{subfigure}[b]{0.47\textwidth}
			\includegraphics[width=\textwidth]{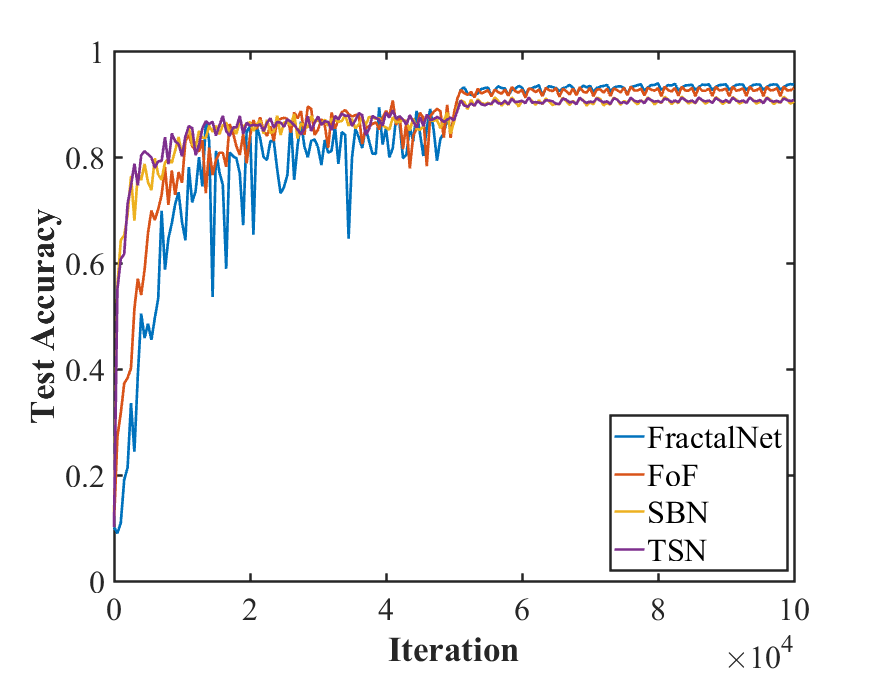}
			\caption{Cifar-10}
			\label{fig:Cifar10FoFSBNTSN}
		\end{subfigure}
		\quad
		\centering
		\begin{subfigure}[b]{0.47\textwidth}
			\includegraphics[width=\textwidth]{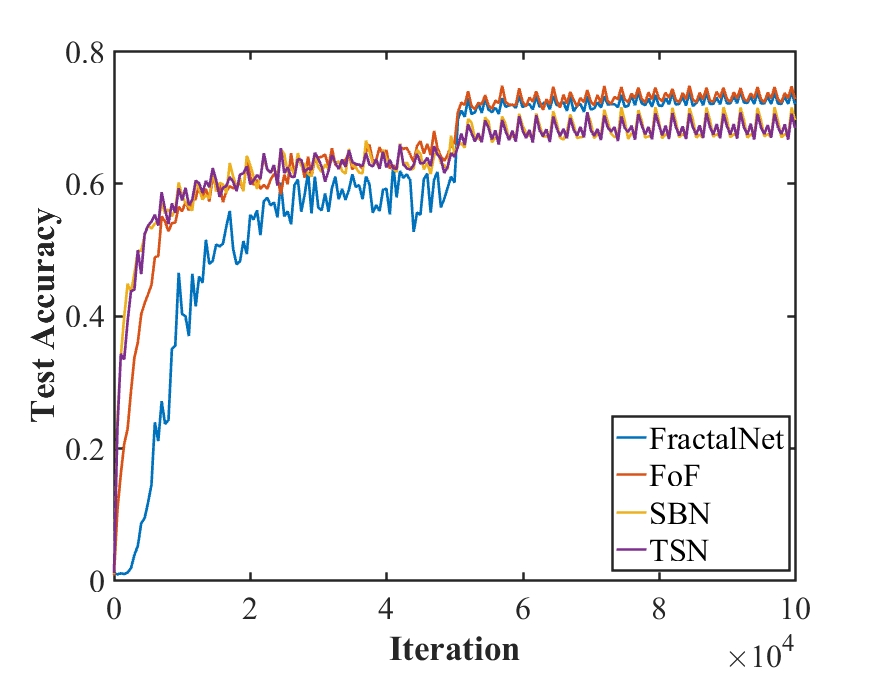}
			\caption{Cifar-100}
			\label{fig:Cifar100FoFSBNTSN}
		\end{subfigure}
		\caption{Test accuracy of the original FractalNet compared with FoF, SBN, and TSN networks.}
		\label{fig:FoFSBNTSNCompare}
	\end{figure}

	\subsection{Results}
	\label{sec:results}
	
	The experiments in this section are primarily to empirically validate the architectural innovations described above but not to fully test them.  We leave a more complete evaluation to future work. 
	
	Table \ref{tab:results} and Figures  \ref{fig:fractalCompare} and \ref{fig:FoFSBNTSNCompare} compare the final test accuracy results for CIFAR-10 and CIFAR-100 in a number of experiments.
	An accuracy value in  Table  \ref{tab:results} is computed as the mean of the last 6 test accuracies computed over the last 3,000 iterations (out of 100,000) of the training.
	The results from the original FractalNet (\citealt{larsson2016fractalnet}) are given in the first row of the table and we use this as our baseline.
	The first four rows of Table \ref{tab:results}  and Figure  \ref{fig:fractalCompare} compare the test accuracy  of the original FractalNet architectures to architectures with a few modifications advocated by design patterns.
	The first modification is to use concatenation instead of fractal-joins at all the downsampling locations in the networks.
	The results for both CIFAR-10 (\ref{fig:Cifar10fractal}) and CIFAR-100 (\ref{fig:Cifar100fractal} indicate that the results are equivalent when concatenation is used instead of fractal-joins at all the downsampling locations in the networks.
	The second experiment was to change the kernel sizes in the first module from 3x3 such that the left most column used a kernel size of 7x7, the second column 5x5, and the third used 3x3.
	The fractal-join for module one was replaced with Maxout.
	The results in Figure  \ref{fig:fractalCompare} are a bit worse, indicating that the more cooperative fractal-join (i.e., mean/summation) with 3x3 kernels has better performance than the competitive Maxout with multiple scales.
	Figure  \ref{fig:fractalCompare} also illustrates how an experiment replacing max pooling with average pooling throughout the architecture changes the training profile.
	For CIFAR-10, the training accuracy rises quickly, plateaus, then lags behind the original FractalNet but ends with a better final performance, which  implies that average pooling might significantly reduce the length of the training (we plan to examine this in future work). 
	This behavior provides some evidence that ``cooperative'' average pooling might be preferable to ``competitive'' max pooling.
	
	Table \ref{tab:results} and Figure \ref{fig:FoFSBNTSNCompare} compare the  test accuracy results for the architectural innovations described in Section \ref{sec:innovations}.
	The FoF architecture ends with a similar final test accuracy as FractalNet but the SBN and TSN architectures (which use freeze-drop-path) lag behind when the learning rate is dropped.
	However, it is clear from both Figures \ref{fig:Cifar10FoFSBNTSN} and \ref{fig:Cifar100FoFSBNTSN} that the new architectures train more quickly than FractalNet.
	The FoF network is best as it trains more quickly than FractalNet and achieves similar accuracy. 
	The use of freeze-drop-path in SBN and TSN is questionable since the final performance lags behind FractalNet, but we are leaving the exploration for more suitable applications of these new architectures for future work.
	
	\section{Conclusion}
	
	In this paper we describe convolutional neural network architectural design patterns that we discovered by studying the  plethora of new architectures in recent deep learning papers.
	We hope these design patterns will be useful to both experienced practitioners looking to push the \SotA and novice practitioners looking to apply deep learning to new applications.
	There exists a large expanse of potential follow up work (some of which we have indicated here as future work).
	Our effort here is primarily focused on Residual Networks for classification but we hope this preliminary work will inspire others to follow up with new architectural design patterns for Recurrent Neural Networks, Deep Reinforcement Learning architectures, and beyond.

	\subsubsection*{Acknowledgments}
	The authors want to thank the numerous researchers upon whose work these design patterns are based and especially  \citealt{larsson2016fractalnet} for making their code publicly available.
	This work was supported by the US Naval Research Laboratory base program.

	\bibliography{CNNdesignPatterns.bib}
	\bibliographystyle{iclr2017_conference}

	\appendix
	
	\section{Relationships between residual architectures}
	\label{sec:equiv}
	
	The architectures mentioned in Section \ref{sec:relwork} commonly combine outputs from two or more layers using concatenation along the depth axis, element-wise summation, and element-wise average. We show here that the latter two are special cases of the former with  weight-sharing enforced. 
	Likewise, we show that skip connections can be considered as introducing additional layers into a network that share parameters with existing layers. In this way, any of the Residual Network variants can be reformulated into a standard form where many of the variants  are equivalent. 
	
	A filter has three dimensions: two spatial dimensions, along which convolution occurs, and a third dimension, depth. Each input channel corresponds to a different depth for each filter of a layer. As a result, a filter can be considered to consist of ``slices,'' each of which is convolved over one input channel. The results of these convolutions are then added together, along with a bias, to produce a single output channel. The output channels of multiple filters are concatenated to produce the output of a single layer. 
	When the outputs of several layers are concatenated, the behavior is similar to that of a single layer. However, instead of each filter having the same spatial dimensions, stride, and padding, each filter may have a different structure. 
	As far as the function within a network, though, the two cases are the same. In fact, a standard layer, one where all filters have the same shape, can be considered a special case of concatenating outputs of multiple layer types. 
	
	If summation is used instead of concatenation, the network can be considered to perform concatenation but enforce weight-sharing in the following layer. The results of first summing several channels element-wise and then convolving a filter slice over the output yields the same result as convolving the slice over the channels and then performing an element-wise summation afterwards. Therefore, enforcing weight-sharing such that the filter slices applied to the $n$th channel of all inputs share weight results in behavior identical to summation, but in a form similar to concatenation, which highlights the relationship between the two. 
	When Batch Normalization (BN) (\citealt{ioffe2015batch} is used, as is the current standard practice, performing an average is essentially identical to performing a summation, since BN scales the output. 
	Therefore, scaling the input by a constant (i.e., averaging instead of a summation) is rendered irrelevant. 
	The details of architecture-specific manipulations of summations and averages is described further in Section \ref{sec:joining}. 
	
	Due to the ability to express depth-wise concatenation, element-wise sum, and element-wise mean as variants of each other, architectural features of recent works can be combined within a single network, regardless of choice of combining operation. However, this is not to say that concatenation has the most expressivity and is therefore strictly better than the others. 
	Summation allows networks to divide up the network's task.
	Also, there is a trade-off between the number of parameters and the expressivity of a layer; summation uses weight-sharing to significantly reduce the number of parameters within a layer at the expense of some amount of expressivity. 
	
	Different architectures can further be expressed in a similar fashion through changes in the connections themselves. A densely connected series of layers can be ``pruned'' to resemble any desired architecture with skip connections through zeroing specific filter slices. This operation removes the dependency of the output on a specific input channel; if this is done for all channels from a given layer, the connection between the two layers is severed. Likewise, densely connected layers can be turned into linearly connected layers while preserving the layer dependencies; a skip connection can be passed through the intermediate layers. A new filter can be introduced for each input channel passing through, where the filter performs the identity operation for the given input channel. All existing filters in the intermediate layers can have zeroed slices for this input so as to not introduce new dependencies. In this way, arbitrarily connected layers can be turned into a standard form. 
	
	We certainly do not recommend this representation for actual experimentation as it introduces fixed parameters. We merely describe it to illustrate the relationship between different architectures. This representation illustrates how skip connections effectively enforce specific weights in intermediate layers. Though this restriction reduces expressivity, the number of stored weights is reduced, the number of computations performed is decreased, and the network might be more easily trainable. 

	\section{Implementation Details}
	\label{sec:impdetails}

	Our implementations are in Caffe (\citealt{jia2014caffe}; downloaded October 9, 2016) using CUDA 8.0.  
	These experiments were run on a 64 node cluster with 8 Nvidia Titan Black GPUs, 128 GB memory, and dual Intel Xenon E5-2620 v2 CPUs per node.  
	We used the CIFAR-10 and CIFAR-100 datasets (\citealt{krizhevsky2009learning} for our classification tests.
	These datasets consist of 60,000 32x32 colour images (50,000 for training and 10,000 for testing) in 10 or 100 classes, respectively.
	Our Caffe code and prototxt files are publicly available at https://github.com/iPhysicist/CNNDesignPatterns.

	\subsection{Architectures}
	
	We started with the FractalNet implementation \footnote{https://github.com/gustavla/fractalnet/tree/master/caffe}
	as our baseline and it is described in \citealt{larsson2016fractalnet}.
	We used the three column module as shown in Figure \ref{fig:Fractmodule}.
	In some of our experiments, we replaced the fractal-join with concatenation at the downsampling locations.
	In other experiments, we modified the kernel sizes in module one and combined the branches with Maxout.
	A FractalNet module is shown in Figure \ref{fig:Fractmodule}  and the architecture consists of five sequential modules.
	
	Our fractal of FractalNet (FoF) architecture uses the same module but has an overall fractal design as in Figure \ref{fig:FoFarch} rather than the original sequential one. 
	We limited our investigation to this one realization and left the study of other (possibly more complex) designs for future work.
	We followed the FractalNet implementation in regards to dropout where the dropout rate for a module were 0\%, 10\%, 20\%, or 30\%, depending on the depth of the module in the architecture.
	This choice for dropout rates were not found by experimentation and better values are possible.
	The local drop-path rate in the fractal-joins were fixed at 15\%, which is identical to the FractalNet implementation.
	
	Freeze-drop-path introduces four new parameters.
	The first is whether the active branch is chosen stochastically or deterministically.
	If it is chosen stochastically, a random number is generated and the active branch is assigned based on which interval it falls in (intervals will be described shortly).
	If it is deterministically, a parameter is set by the user as to the number of iterations in one cycle through all the branches (we called this parameter $ num\_iter\_per\_cycle$).
	In our Caffe implementation of the freeze-drop-path unit, the bottom input specified first is assigned as branch 1, the next is branch 2, then branch 3, etc.
	The next parameter indicates the proportion of iterations each branch should be active relative to all the other branches.
	The first type of interval uses the square of the branch number (i.e., $ 1, 4, 9, 16, ...$) to assign the interval length for that branch to be active, which gives the more update iterations to the higher numbered branches.
	The next type gives the same amount of iterations to each branch.
	Our experiments showed that the first interval type works better (as we expected) and was used to obtained the results in Section \ref{sec:results}.
	In addition, our experiments showed that the stochastic option works better than the deterministic option (unexpected) and was used for Section \ref{sec:results} results.
	
	The Stagewise Boosting Network's (SBN) architecture is the same as the FoF architecture except that branches 2 and 3 are combined with a fractal-join and then combined with branch 1 in a freeze-drop-path join. 
	The reason for combining branches 2 and 3 came out of our first experiments; if branches 2 and 3 were separate, the performance deteriorated when branch 2 was frozen and branch 3 was active.  
	In hindsight, this is due to the weights in the branch 2 path that are also in branch 3's path being modified by the training of branch 3.
	The Taylor series network has the same architecture as SBN with the addition of squaring the branch 2 and 3 combined activations before the freeze-drop-path join.
	
	For all of our experiments, we trained for 400 epochs.
	Since the training  used 8 GPUs and each GPU had a batchsize of 25, 400 epochs amounted to 100,000 iterations.
	We adopted the same learning rate as the FractalNet implementation, which started at 0.002 and dropped the learning rate by a factor of 10 at epochs 200, 300, and 350.

\end{document}